\documentclass[10pt,letterpaper,twocolumn]{article}
\usepackage[T1]{fontenc}
\usepackage[margin=0.75in,columnsep=0.3in]{geometry}
\usepackage{newtxtext}
\usepackage{helvet}
\usepackage{courier}
\usepackage{graphicx}
\usepackage[hyphens]{url}
\usepackage[authoryear,round]{natbib}
\usepackage[font=small,labelfont=bf]{caption}
\usepackage{algorithm}
\usepackage{algpseudocode}
\usepackage{newfloat}
\usepackage{listings}
\usepackage{booktabs}
\usepackage{amsmath}
\usepackage{amssymb}
\usepackage{multirow}
\usepackage{placeins}
\usepackage{microtype}
\usepackage[hidelinks]{hyperref}

\DeclareCaptionStyle{ruled}{labelfont=normalfont,labelsep=colon,strut=off}
\floatstyle{ruled}
\newfloat{listing}{tb}{lst}
\floatname{listing}{Listing}

\hypersetup{
  pdftitle={From Outcomes to Strategies: Learning Strategy Utility for Mathematical Reasoning},
  pdfauthor={Ruikang Zhang, Xiao An, Xuli Shen, Jiaxing Sun, Xiaoyi Yu, Jin Zeng, Jiang Wu, Tong Lin},
  pdfsubject={Strategy utility learning for mathematical reasoning},
  pdfkeywords={mathematical reasoning, strategy utility, reinforcement learning, reward models}
}

\title{\LARGE\bfseries From Outcomes to Strategies: Learning Strategy Utility\\
for Mathematical Reasoning}
\author{
  \normalsize \textbf{Ruikang Zhang}\textsuperscript{1,}\thanks{These authors contributed equally.}\quad
  \textbf{Xiao An}\textsuperscript{2,*}\quad
  \textbf{Xuli Shen}\textsuperscript{3}\quad
  \textbf{Jiaxing Sun}\textsuperscript{3}\\[0.25em]
  \normalsize \textbf{Xiaoyi Yu}\textsuperscript{4}\quad
  \textbf{Jin Zeng}\textsuperscript{5}\quad
  \textbf{Jiang Wu}\textsuperscript{3,}\thanks{Corresponding authors.}\quad
  \textbf{Tong Lin}\textsuperscript{1,\textdagger}\\[0.6em]
  \small \textsuperscript{1}Peking University\quad
  \textsuperscript{2}Wuhan University\\
  \small \textsuperscript{3}Shanghai Artificial Intelligence Laboratory\\
  \small \textsuperscript{4}Renmin University of China\quad
  \textsuperscript{5}Tongji University
}
\date{}

\begin{document}
\maketitle

\begin{abstract}
Reinforcement learning with verifiable rewards has substantially improved mathematical reasoning. However, terminal correctness alone provides limited insight into the quality of high-level strategies, such as theorem selection and subgoal decomposition, when considered separately from their subsequent execution. This paper studies strategy utility, which is defined as the likelihood that a strategy supports a correct downstream solution under a given executor. We introduce SURE, a framework for learning and leveraging relative strategy utility. In this framework, high-level strategies are separated from their detailed reasoning. Based on the pairwise preferences constructed from strategy-conditioned rollouts and teacher-generated contrasts, a Strategy Reward Model is learned to estimate relative strategy utility. During reinforcement learning, the frozen reward model reads only the extracted strategy, whose score is combined with the correctness and format rewards in a sequence-level GRPO objective. Compared with outcome-and-format GRPO baselines, experiments show that SURE improves average pass@1 by 1.87\%, 2.64\%, and 2.93\% across three policy backbones. Our method also achieves competitive or better accuracy than stronger reward baselines while requiring substantially lower GRPO-stage compute.
\end{abstract}

\section{Introduction}
\label{sec:introduction}
Reinforcement learning with verifiable rewards has become a scalable approach for improving mathematical reasoning in large language models~\citep{deepseekai2025deepseekr1}.
Group relative policy optimization (GRPO)~\citep{shao2024deepseekmath} estimates advantages from grouped rollouts and supports large-scale training on mathematical tasks such as GSM8K~\citep{cobbe2021training} and MATH~\citep{hendrycks2021math}.
\begin{figure}[!h]
\centering
\includegraphics[width=\linewidth]{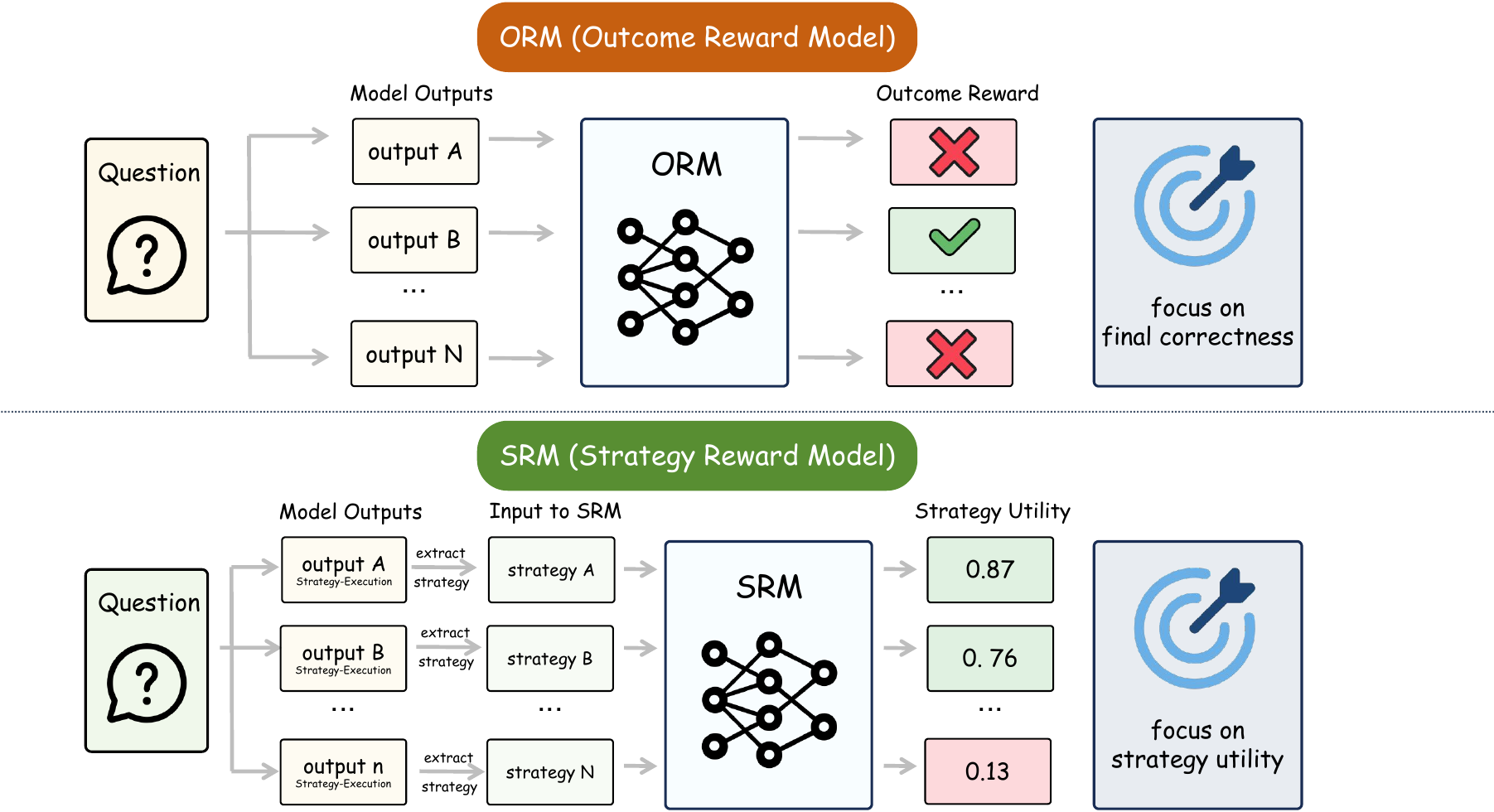}
\caption{\textbf{Outcome rewards and strategy utility signals.} An outcome reward evaluates the final answer. SURE extracts the strategy and uses the SRM to estimate its relative utility before forming the sequence-level training reward.}
\label{fig:teaser}
\end{figure}
However, terminal correctness remains the dominant training signal.
It reveals whether a complete response succeeds, but it does not identify which decisions made that success probable.
A mathematical response contains two distinct types of decisions.
The strategy chooses the method, relevant theorems, intermediate targets, and order of subgoals~\citep{schoenfeld1981episodes}.
The execution applies those choices through concrete calculations and derivations.
A terminal reward observes only the final answer and therefore conflates strategy selection with detailed execution.
A sound strategy can fail because of an execution error, while a fragile strategy can succeed through an unusual derivation or a lucky correction.
In both cases, the final outcome provides limited guidance about which solution route the policy should learn.
Figure~\ref{fig:teaser} illustrates this distinction.

We study \emph{strategy utility} to provide supervision at this missing level.
For a given problem, executor, and decoding configuration, strategy utility is the probability that the strategy supports a correct solution.
Strategy utility is conditioned on a specified reference executor and decoding configuration, rather than being an intrinsic property of the strategy text.
Accordingly, the learned SRM estimates relative utility under this reference setting and is used in downstream optimization as a potentially transferable ranking signal, rather than as a universal estimator valid for all downstream executors.
Strategy utility lies between outcome supervision, which evaluates complete solutions~\citep{cobbe2021training,ouyang2022training}, and process supervision, which evaluates individual reasoning steps~\citep{lightman2023let,wang2024mathshepherd}.
Prior work has introduced explicit intermediate structure through chain-of-thought prompting, decomposition, plan-and-solve templates, and inference-time search~\citep{wei2022chain,wang2023plan,yao2023tree}.
These methods establish the value of planning, but they generally use it as a prompting format or an inference procedure.
We instead ask whether relative strategy utility can be learned and used as a training signal.

We introduce \textbf{SURE}, a framework for learning \textbf{S}trategy \textbf{U}tility for mathematical \textbf{RE}asoning.
SURE first uses supervised fine-tuning to teach a policy to emit a high-level strategy before detailed reasoning.
This structure makes the strategy explicit and allows it to be evaluated separately from its execution.
SURE then constructs pairwise strategy preferences from two data sources.
Strategy-conditioned rollouts compare candidate strategies through their empirical execution success, with confidence filtering to reduce noise from finite sampling.
Teacher-generated contrasts add strategies with explicit structural errors that may not appear in policy rollouts.
These preferences train a Strategy Reward Model (SRM) to estimate relative utility from the problem and strategy alone.
The SRM never observes the execution or final answer when assigning a score.

\begin{figure*}[!t]
\centering
\includegraphics[width=0.95\linewidth]{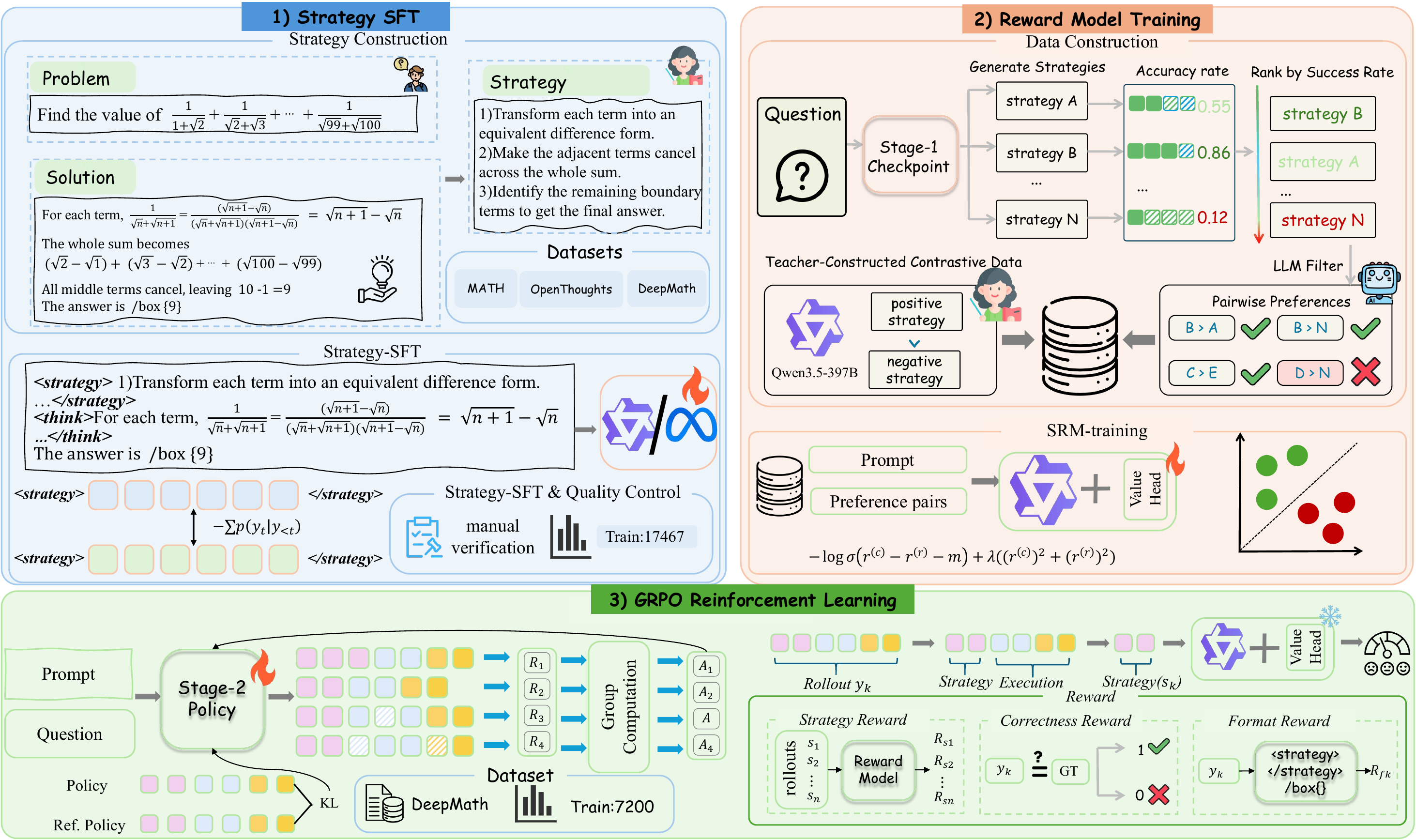}
\caption{Overview of SURE. \textbf{1 SFT} teaches the policy to separate strategy from execution. \textbf{2 SRM training} constructs pairwise preferences and learns relative strategy utility. \textbf{3 Reinforcement learning} combines the frozen SRM score with correctness and format rewards. We use GRPO for policy optimization in our experiments.}
\label{fig:pipeline}

\end{figure*}

During reinforcement learning, SURE freezes the SRM and uses its score as an auxiliary reward.
In our GRPO instantiation, the score is combined with correctness and format rewards in a sequence-level objective.
The current design normalizes SRM scores within each rollout group and suppresses the utility reward for incorrect responses.
These operations preserve terminal correctness as the primary objective and reduce incentives to exploit the learned reward.
SURE defines the strategy utility signal and its use during training, while GRPO is the policy optimizer used in our experiments.
Figure~\ref{fig:pipeline} summarizes the complete framework.

We organize our evaluation around three questions.
\begin{enumerate}
\item[\textbf{RQ1}] Does the learned SRM rank strategies consistently under a fixed executor?
\item[\textbf{RQ2}] Does SURE improve policy optimization beyond matched outcome and format rewards, and can it be combined with process rewards?
\item[\textbf{RQ3}] Given a pretrained SRM, how does the GRPO-stage cost compare with online plan reward and process reward baselines~\citep{dou2025plan}?
\end{enumerate}

In summary, our main contributions are:
\begin{itemize}
\item We formulate relative strategy utility as an intermediate supervision target for mathematical reasoning. The target measures how likely a strategy is to support a correct solution under a specified executor and decoding configuration.
\item We introduce \textbf{SURE}, which learns relative utility from pairwise strategy preferences and uses the learned signal as an auxiliary reward for policy optimization.
\item Across three policy backbones, SURE improves average pass@1 over matched outcome and format GRPO baselines by $1.87\%$, $2.64\%$, and $2.93\%$. We also evaluate process-reward integration, controlled SFT baselines, SRM data-source and reward-design ablations, and GRPO-stage compute, demonstrating that learned strategy utility is an effective and computationally practical training signal for mathematical reasoning.
\end{itemize}

\section{Related Work}
\label{sec:related}
\subsection{Explicit Plans and Decompositions}

Chain-of-thought prompting and self-consistency show that explicit intermediate reasoning and multi-path aggregation can improve mathematical problem solving~\citep{wei2022chain,wang2023selfconsistency}.
Decomposition methods~\citep{zhou2022least,khot2022decomposed} and plan-and-solve templates~\citep{wang2023plan} make high-level decisions visible before detailed derivations.
Tree- and graph-based methods further use these decisions to guide inference-time search~\citep{yao2023tree,besta2024graph,hao2023reasoning}.
This line of work establishes the value of planning, but usually treats a plan as a prompt format or a search object.
SURE instead treats the strategy as an explicit learning target and estimates its relative utility for policy optimization.

\subsection{Reward Models for Mathematical Reasoning}

Outcome reward models score complete solutions and can provide terminal supervision for reinforcement learning~\citep{cobbe2021training,ouyang2022training}.
Process reward models provide denser feedback by evaluating individual reasoning steps~\citep{lightman2023let,uesato2022solving,rezaei2026progrs,yue2026promoting,xu2025logicreward,yuan2026verifiable,chen2026structured}.
Automated labels can reduce the cost of process supervision~\citep{wang2024mathshepherd,wang2025masked}, but label noise and reward misalignment remain important concerns~\citep{zhang2025lessons,cui2025prime}.
These reward models operate at the level of a full solution or a local step.
SURE introduces an intermediate target, \emph{strategy utility}, that focuses on the high-level method, theorem choices, and subgoal structure.
Its training preferences use execution evidence, while the resulting Strategy Reward Model scores only the problem and the extracted strategy.
This separation gives policy optimization a strategic signal without asking the reward model to reproduce detailed derivations.

\subsection{Reinforcement Learning with Planning-Aware Rewards}

GRPO and related group-normalized policy optimization methods estimate advantages from grouped rollouts and have become common for mathematical reasoning~\citep{shao2024deepseekmath,deepseekai2025deepseekr1,wen2025rlvr,he2025random,zhang2025r1vl}.
In these settings, answer-level rewards provide a reliable correctness criterion but offer limited information about early strategic decisions.
Planning-aware RL methods address this limitation by evaluating plan quality during training, often through additional online executions~\citep{dou2025plan}.
Such evaluations are tied to the current policy and must be repeated as the policy changes.

SURE learns relative strategy utility offline from pairwise preferences and reuses a frozen Strategy Reward Model during policy optimization.
The SRM score is computed from the problem and extracted strategy, not from the execution tokens of an individual rollout.
It can therefore be combined with correctness and format rewards in a sequence-level objective.
Although we instantiate policy optimization with GRPO in our experiments, the definition of reference-conditioned strategy utility and the training of the SRM do not inherently depend on GRPO.

\section{Methodology}
\label{sec:method}
We present \textbf{SURE}, a three-stage framework for learning and using relative strategy utility in mathematical reasoning.
Figure~\ref{fig:pipeline} provides a schematic overview.
We first formalize the strategy and execution decomposition, then describe strategy-guided SFT, SRM data construction and training, and policy optimization with SURE.

\subsection{Strategy and Execution Decomposition}
\label{sec:decomposition}

For a mathematical problem $x$, we factor the model response into a \textbf{strategy} $s$ and an \textbf{execution} $y$.
The strategy is a high-level solution outline that specifies tool choice, intermediate targets, and subgoal ordering.
The execution is the concrete derivation that produces the final answer.
The strategy excludes detailed computation and the final answer.

Let $q(x,s)$ denote the strategy utility under the fixed executor and decoding configuration used for evaluation.
We define it as the probability that the executor produces a correct solution after conditioning on the strategy.
\begin{equation}
\label{eq:strategy-utility}
q(x, s) = P(\text{correct} \mid x, s).
\end{equation}
This definition evaluates a strategy through its expected usefulness rather than its surface plausibility.
The decomposition is enforced by structured output tags \texttt{<strategy>}~\texttt{</strategy>} and \texttt{<think>}~\texttt{</think>}.
The tags are taught during SFT and checked during policy optimization.
Figure~\ref{fig:teaser} contrasts this construction with an outcome reward model.
An ORM labels a complete generation, whereas the SRM scores the extracted strategy and ranks alternative routes for the same problem.

\subsection{Strategy-Guided SFT}
\label{sec:sft}

The first stage establishes the strategy and execution format through supervised fine-tuning.
We draw problems from the MATH training split, OpenThoughts, and DeepMath.
For each problem, we prompt Qwen3.5-397B-A17B with the reference solution and ask it to generate a high-level strategy without intermediate computations.
We manually verify the resulting question, strategy, and execution triples using the SFT prompt in the appendix.
The verified SFT corpus contains $17{,}467$ examples.
Each example concatenates the teacher strategy and the reference execution under the structured tags.
The policy is trained with standard next-token prediction over the full sequence.
This stage establishes a reliable output structure for the subsequent preference learning and policy optimization stages.

\subsection{SRM Training Data Construction}
\label{sec:srm-data}

The SRM is trained on pairwise preferences $(x, s^{+}, s^{-})$ in which $s^{+}$ has higher empirical execution success than $s^{-}$.
We construct these preferences from two sources using $32{,}000$ problems sampled from DeepMath and OpenThoughts.

\subsubsection{Rollout-Induced Preference Data}

For each training problem $x$, we sample $K = 8$ candidate strategies $\{s^{(k)}\}_{k=1}^{K}$ from the Qwen3-4B Stage-1 policy $\pi_{\text{SFT}}(\cdot \mid x)$ with temperature $T_s = 0.95$, where $k$ indexes the candidate strategy.
For each candidate $s^{(k)}$, we then sample $M = 24$ independent execution rollouts $y^{(k,j)} \sim \pi_{\text{SFT}}(\cdot \mid x, s^{(k)})$, $j=1,\ldots,M$, while holding the strategy prefix fixed.
This procedure yields $KM = 192$ execution rollouts per problem.
Let $\ell^{(k,j)}$ denote the binary correctness label assigned by the answer verifier.
The empirical success rate of candidate strategy $s^{(k)}$ is
\begin{equation}
\label{eq:empirical-rate}
\begin{aligned}
\ell^{(k,j)} &= \mathbf{1}\!\left[\operatorname{Verify}\!\left(y^{(k,j)}\right)\right], \\
\hat{q}(x, s^{(k)}) &= \frac{1}{M} \sum_{j=1}^{M} \ell^{(k,j)}.
\end{aligned}
\end{equation}

We construct pairwise preferences with confidence-aware filtering.
We model each execution outcome as Bernoulli with probability $q(x,s)$.
We retain a pair $(s^{(i)} \succ s^{(j)})$ when the posterior probability that $s^{(i)}$ has higher utility than $s^{(j)}$ exceeds $0.9$.
\begin{equation}
\label{eq:confidence}
\mathbb{P}\bigl[q(x, s^{(i)}) > q(x, s^{(j)})\bigr] \geq 0.9,
\end{equation}
We compute this probability with a Beta-Binomial model and a $\text{Beta}(1,1)$ prior.
The filter removes pairs for which the empirical difference can plausibly be sampling noise~\citep{wang2024mathshepherd,zhang2025lessons}.
After filtering, $53{,}464$ candidate pairs remain.
We then use Qwen3.5-397B-A17B for preference verification~\citep{zheng2023llmjudge}.
The judge receives the problem and the proposed pair $(s^{+},s^{-})$ and decides whether the labeled preference is reasonable.
Unreasonable pairs are discarded, leaving $48{,}413$ pairs in $\mathcal{D}_{\text{rollout}}$.
The prompt template appears in the preference-verification appendix.

\subsubsection{Teacher-Constructed Contrastive Data}

Rollout-induced preferences are limited to strategies that the SFT policy can generate.
We therefore also use Qwen3.5-397B-A17B to construct positive strategies $s^{+}$ and negative strategies $s^{-}$ for each question.
Positive strategies specify sound theorems and feasible subgoal orderings.
Negative strategies contain flawed, incomplete, or misleading approaches.
The resulting $17{,}197$ pairs form $\mathcal{D}_{\text{teacher}}$.
The prompt template appears in the teacher-contrast appendix.

\subsubsection{Merging and Filtering}

We merge $\mathcal{D}_{\text{rollout}}$ and $\mathcal{D}_{\text{teacher}}$ into
$\mathcal{D}_{\text{SRM}} = \{(x_i, s_i^{+}, s_i^{-})\}_{i=1}^{N_{\text{SRM}}}$.
The merged set contains $65{,}610$ preference pairs in total.
Full statistics are provided in the appendix.
We balance the two sources during training to reduce sensitivity to rollout noise and teacher-specific phrasing.

\subsection{SRM Training}
\label{sec:srm-training}

The SRM $r_\phi(x,s)$ is initialized from Qwen2.5-7B-Instruct~\citep{yang2024qwen25} and adapted with LoRA of rank 64~\citep{hu2022lora}.
A scalar value head over the final strategy token's hidden state replaces the language-model head.
The SRM receives the problem and strategy, but never the execution or final answer.

We train the SRM with a margin-augmented Bradley--Terry pairwise ranking loss~\citep{bradley1952rank} and a scale-anchoring regularizer~\citep{christiano2017deep}.
\begin{equation}
\label{eq:srm-loss}
\begin{aligned}
\mathcal{L}_{\text{SRM}}(\phi) ={}&
-\log \sigma\!\bigl(r^{(c)} - r^{(r)} - m\bigr) \\
&+ \lambda \left(\bigl(r^{(c)}\bigr)^2 + \bigl(r^{(r)}\bigr)^2\right),
\end{aligned}
\end{equation}
where $r^{(c)} \!=\! r_\phi(x, s^{+})$ and $r^{(r)} \!=\! r_\phi(x, s^{-})$ are the preferred and dispreferred scores.
The margin $m \!\geq\! 0$ enforces a minimum score gap.
Rollout-induced pairs use weights derived from the confidence score in Equation~\ref{eq:confidence}, while teacher pairs use uniform weights.
The trained SRM is frozen before policy optimization.
We validate its ranking behavior with a fixed-executor utility ranking study.

\subsection{SURE Policy Optimization}
\label{sec:sure-rl}

The third stage uses the frozen SRM during policy optimization.
Our experiments instantiate this stage with GRPO~\citep{shao2024deepseekmath}.
Each response receives a composite sequence-level reward
\begin{equation}
\label{eq:composite-reward}
R = \alpha \, R_{\text{correct}} + \beta \, R_{\text{format}} + \gamma \, R_{\text{SRM}},
\end{equation}
where $\alpha$, $\beta$, and $\gamma$ are scalar weights for the correctness, format, and strategy utility terms, respectively. The component rewards satisfy $R_{\text{correct}} \in \{0, 1\}$ for answer correctness, $R_{\text{format}} \in \{0, 1\}$ for the structured output format, and $R_{\text{SRM}} = r_\phi(x,s)$ for the extracted strategy.

At scoring time, the SRM receives only the problem and strategy.
It does not observe the execution tokens or the final answer.
The scalar SRM score is combined with correctness and format rewards at the sequence level.
Unlike process reward models (PRMs), which evaluate local reasoning steps~\citep{lightman2023let,wang2024mathshepherd}, the SRM evaluates the selected high-level approach using only the problem and extracted strategy.
It therefore provides strategy-level guidance without scoring the execution trace or final answer.

We use two stabilizers motivated by reward design analyses~\citep{gao2024designing,cui2025prime}.
First, group-wise normalization maps SRM scores to zero mean and unit variance within each GRPO group.
Second, correctness gating defines
\begin{equation}
\label{eq:gated-srm}
R_{\text{SRM}}^{\text{gated}} = R_{\text{correct}} \cdot R_{\text{SRM}}.
\end{equation}
The gate sets the utility contribution to zero for incorrect responses.
Because the SRM is an approximation of strategy utility, this constraint limits incentives to exploit discrepancies between the learned score and verified correctness.
This gating is designed to reduce reward hacking by preventing an incorrect response from receiving a positive strategy utility contribution.
The utility score therefore refines the ranking of correct responses rather than replacing correctness as the primary objective.

In our experiments, GRPO uses the composite reward with group-based advantage normalization and KL regularization toward $\pi_{\text{SFT}}$~\citep{shao2024deepseekmath}.
RL is run on $7{,}200$ problems sampled from DeepMath, with the SRM frozen throughout.
Unlike methods that estimate plan quality online during RL~\citep{dou2025plan}, SURE estimates utility once offline and reuses the learned estimator throughout RL.
The GRPO-stage cost analysis treats the SRM as a reusable pretrained reward model.

\section{Experiments}
\label{sec:experiments}
\begin{table*}[!t]
\centering
\small
\caption{Main results on pass@1 (\%) across four mathematical reasoning benchmarks. \textbf{Bold} marks the best result within each backbone block. Both GRPO initialization controls use outcome and format rewards.}
\label{tab:main}
\setlength{\tabcolsep}{4pt}
\resizebox{0.8\linewidth}{!}{%
\begin{tabular}{llcccc|c}
\toprule
Backbone & Method & AIME24 $\uparrow$ & AIME25 $\uparrow$ & AMC23 $\uparrow$ & MATH500 $\uparrow$ & Avg $\uparrow$ \\
\midrule
\multirow{7}{*}{Qwen3-4B}
 & Base                              & 23.00 & 22.67 & 64.00 & 76.76 & 46.61 \\
 & CoT-SFT (token-matched)           & 36.67 & 24.33 & 76.50 & 77.88 & 53.85 \\
 & SFT (strategy-guided)             & 42.00 & 29.67 & 80.00 & 81.82 & 58.37 \\
 & GRPO from CoT-SFT                 & 38.33 & 26.67 & 78.00 & 78.96 & 55.49 \\
 & GRPO from Strategy-SFT            & 40.67 & 30.00 & 80.50 & 82.32 & 58.37 \\
 & PTA-GRPO~\citep{dou2025plan}      & 40.33 & 31.00 & 82.25 & 81.54 & 58.78 \\
 & \textbf{SURE} (ours)              & \textbf{42.67} & \textbf{31.00} & \textbf{84.25} & \textbf{83.02} & \textbf{60.24} \\
\midrule
\multirow{7}{*}{Qwen3-8B}
 & Base                              & 26.00 & 21.00 & 62.75 & 78.20 & 46.99 \\
 & CoT-SFT (token-matched)           & 37.33 & 23.67 & 77.25 & 78.24 & 54.12 \\
 & SFT (strategy-guided)             & 41.33 & 31.33 & 82.75 & 82.02 & 59.36 \\
 & GRPO from CoT-SFT                 & 38.00 & 24.67 & 77.75 & 78.88 & 54.83 \\
 & GRPO from Strategy-SFT            & 39.33 & 31.00 & 79.25 & 82.10 & 57.92 \\
 & PTA-GRPO~\citep{dou2025plan}      & 39.67 & 31.33 & 82.50 & 81.94 & 58.86 \\
 & \textbf{SURE} (ours)              & \textbf{42.33} & \textbf{32.67} & \textbf{84.50} & \textbf{82.72} & \textbf{60.56} \\
\midrule
\multirow{7}{*}{Llama-3.2-3B-Instruct}
 & Base                              & 3.33  & 3.33  & 19.50 & 39.40 & 16.39 \\
 & CoT-SFT (token-matched)           & 2.67  & 1.33  & 21.00 & 43.20 & 17.05 \\
 & SFT (strategy-guided)             & 2.33  & 1.00  & 21.25 & 44.36 & 17.24 \\
 & GRPO from CoT-SFT                 & 3.00  & 1.33  & 21.25 & 44.10 & 17.42 \\
 & GRPO from Strategy-SFT            & 3.00  & 2.00  & 25.00 & 46.60 & 19.15 \\
 & PTA-GRPO~\citep{dou2025plan}      & 3.33  & 2.33  & 25.25 & 46.66 & 19.39 \\
 & \textbf{SURE} (ours)              & \textbf{6.67} & \textbf{3.33} & \textbf{30.00} & \textbf{48.32} & \textbf{22.08} \\
\bottomrule
\end{tabular}%
}

\end{table*}

We evaluate SURE on competition mathematics benchmarks across three policy backbones.
We compare it with outcome-only GRPO, process-reward integration, and online plan-reward estimation.
The experiments address three questions introduced in the Introduction.
RQ1 asks whether SRM scores order strategies consistently under a frozen executor.
RQ2 asks whether the learned utility signal improves policy optimization beyond outcome and format rewards.
RQ3 asks how a reusable pretrained SRM compares with online plan-reward methods in GRPO-stage compute.

\subsection{Experimental Setup}
\label{sec:exp-setup}

\paragraph{Models.}
We instantiate the policy on three backbones spanning two families and a 2.5$\times$ parameter range.
The backbones are Qwen3-4B, Qwen3-8B~\citep{yang2025qwen3}, and Llama-3.2-3B-Instruct.
The strategy reward model is initialized from Qwen2.5-7B-Instruct~\citep{yang2024qwen25} and adapted with LoRA (rank 64)~\citep{hu2022lora}.
All teacher-side strategy generation and judge filtering use Qwen3.5-397B-A17B.

\paragraph{Training data.}
Strategy-guided SFT uses $17{,}467$ manually verified (question, teacher-strategy, reference-execution) triples drawn from MATH~\citep{hendrycks2021math}, OpenThoughts~\citep{guha2025openthoughts}, and DeepMath~\citep{he2025deepmath}.
The SRM is trained on the merged preference set $\mathcal{D}_{\text{SRM}}$.
GRPO is run on a $7{,}200$-problem subset of DeepMath with group size $G\!=\!8$, KL-regularized toward the SFT policy following~\citet{shao2024deepseekmath}.

\paragraph{Benchmarks.}
We report pass@1 on four mathematical benchmarks of varying difficulty.
The benchmarks are AIME 2024, AIME 2025, AMC 2023, and MATH500.
The \emph{Avg} column is the arithmetic mean over the four benchmarks.
For each problem we draw 10 samples at temperature $0.6$ and report the mean pass@1.
Evaluation variance is captured through the 10 inference samples per problem.
Our SFT data draws from the MATH training split, OpenThoughts, and DeepMath. MATH500 is the MATH test split and is disjoint from our training data.
The PRM is the publicly released Qwen2.5-Math-PRM-7B from~\citet{zhang2025lessons,yang2024qwen25math}, kept identical across PRM-only and PRM+SRM runs.

\paragraph{Compared methods.}
We organize the comparisons in three ways.
First, we vary the SFT and RL initialization by comparing Base, CoT-SFT, Strategy-SFT, and GRPO from each SFT checkpoint.
Second, we vary the reward composition by comparing GRPO with outcome and format rewards, with SRM, with PRM, and with both PRM and SRM.
Third, we include a token-matched CoT-SFT control that distills the same teacher into an unstructured format under a matched output-token budget, together with PTA-GRPO~\citep{dou2025plan}, which estimates plan quality online during RL.


\subsection{Main Results (RQ2)}
\label{sec:main-results}

Table~\ref{tab:main} reports pass@1 across the four benchmarks and three backbones.
The main comparison isolates three sources of gain.

\paragraph{Outcome-only GRPO provides limited gains on CoT-SFT.}
Starting from CoT-SFT, outcome and format rewards improve the average by $+1.64\%$, $+0.71\%$, and $+0.37\%$ on Qwen3-4B, Qwen3-8B, and Llama-3.2-3B-Instruct, respectively.
These results show that ordinary CoT policy optimization is useful but modest.

\paragraph{Explicit strategy structure improves outcome-only GRPO.}
The comparison between GRPO from CoT-SFT and GRPO from Strategy-SFT gives gains of $+2.88\%$, $+3.09\%$, and $+1.73\%$ across the three backbones.
Thus, the benefit of the explicit strategy and execution format is not explained by applying outcome-based GRPO to an ordinary CoT format.
The token-matched SFT comparison supports the same conclusion before RL, with Strategy-SFT exceeding CoT-SFT by $+4.52\%$, $+5.24\%$, and $+0.19\%$.

\paragraph{SURE adds a learned utility signal after Strategy-SFT.}
With the same Strategy-SFT initialization, SURE improves over GRPO from Strategy-SFT by $+1.87\%$, $+2.64\%$, and $+2.93\%$.
On Qwen3-8B, outcome and format rewards alone underperform the SFT checkpoint on average (57.92\% vs.\ 59.36\%), whereas SURE reaches 60.56\%.
The consistent gains across backbones support an independent contribution from the learned strategy utility signal.

\paragraph{Online plan rewards provide a weaker reference point.}
PTA-GRPO~\citep{dou2025plan} remains within $\pm 1\%$ of vanilla GRPO on all three backbones and stays $1.5\%$--$2.7\%$ below SURE.
On AMC23, the Qwen3-4B scores are 80.50\% for GRPO, 82.25\% for PTA-GRPO, and 84.25\% for SURE.
This comparison is consistent with SURE's offline precomputation design, since offline construction can use more rollouts per problem and strategy than an online estimate.
The following GRPO-stage cost analysis examines the corresponding cost.

\subsection{SFT Teacher Strength Ablation}
\label{sec:teacher-ablation}

We next vary the teacher used only to construct SFT strategies on Qwen3-4B.
The strong teacher is Qwen3.5-397B-A17B, and the ablation teacher is Qwen3-4B.
The SRM preference data and its judge remain unchanged and still use the strong teacher.
This control therefore measures the sensitivity of SURE to SFT strategy quality rather than teacher independence across the full pipeline.

\begin{table}[t]
\centering
\small
\caption{SFT strategy teacher ablation on Qwen3-4B. The weak-teacher setting changes only the teacher used for SFT strategy construction.}
\label{tab:teacher-ablation}
\setlength{\tabcolsep}{3pt}
\resizebox{\columnwidth}{!}{%
\begin{tabular}{lcccc|c}
\toprule
SFT strategy teacher & AIME24 & AIME25 & AMC23 & MATH500 & Avg \\
\midrule
Strong teacher & 42.67 & 31.00 & \textbf{84.25} & \textbf{83.02} & \textbf{60.24} \\
Qwen3-4B & 42.67 & 31.00 & 82.25 & 82.70 & 59.66 \\
$\Delta$ & 0.00 & 0.00 & $-2.00$ & $-0.32$ & $-0.58$ \\
\bottomrule
\end{tabular}}

\end{table}

Replacing the strong SFT strategy teacher with Qwen3-4B reduces the average by only $0.58\%$.
The decrease is concentrated on AMC23 and MATH500, while AIME24 and AIME25 remain unchanged.
This result suggests that the final gain is not explained entirely by strategy distillation from the largest teacher during SFT.
The exact teacher replacement and the unchanged SRM construction are detailed in the teacher-strength appendix.

\subsection{Compatibility with Process Rewards}
\label{sec:prm-compat}

A natural question is how the SRM signal relates to step-level process rewards.
We compare SURE with (i) GRPO using the publicly released Qwen2.5-Math-PRM-7B from~\citet{zhang2025lessons} and (ii) GRPO using both reward models (Table~\ref{tab:prm}).

\begin{table}[t]
\centering
\small
\caption{PRM and SRM results on Qwen3 backbones with Qwen2.5-Math-PRM-7B. The table reports each reward model separately and their combination.}
\label{tab:prm}
\setlength{\tabcolsep}{3pt}
\resizebox{\columnwidth}{!}{
\begin{tabular}{llcccc|c}
\toprule
Backbone & Method & AIME24 & AIME25 & AMC23 & MATH500 & Avg \\
\midrule
\multirow{3}{*}{Qwen3-4B}
 & SURE (ours)            & 42.67 & 31.00 & 84.25 & \textbf{83.02} & 60.24 \\
 & GRPO + PRM             & 43.00 & \textbf{34.33} & 82.00 & 82.34 & 60.42 \\
 & GRPO + PRM + SRM (ours) & \textbf{45.67} & 33.33 & \textbf{86.00} & 82.70 & \textbf{61.93} \\
\midrule
\multirow{3}{*}{Qwen3-8B}
 & SURE (ours)            & 42.33 & 32.67 & \textbf{84.50} & 82.72 & 60.56 \\
 & GRPO + PRM             & 44.33 & 30.33 & 82.50 & 82.24 & 59.85 \\
 & GRPO + PRM + SRM (ours) & \textbf{45.33} & \textbf{33.33} & 83.50 & \textbf{83.08} & \textbf{61.31} \\
\bottomrule
\end{tabular}}

\end{table}

\paragraph{PRM and SRM provide distinct signals.}
PRM-only and SRM-only achieve comparable averages, with PRM slightly stronger on the smaller backbone and SRM stronger on the larger one.
The averages are $60.42$ vs.\ $60.24$ on Qwen3-4B and $59.85$ vs.\ $60.56$ on Qwen3-8B.
Their per-benchmark profiles also differ.
PRM-only is stronger on AIME25 (Qwen3-4B 34.33), where local-step verification helps catch arithmetic mistakes. SRM-only is stronger on AMC23 (Qwen3-4B 84.25, Qwen3-8B 84.50), where the gain comes from selecting a viable solution route.
Using both signals yields the highest average on both Qwen3 backbones (Qwen3-4B $61.93$, Qwen3-8B $61.31$) and the best AIME24 result overall (45.67, 45.33).

\subsection{Do SRM Scores Track Relative Strategy Utility?}
\label{sec:strategy-effectiveness}

To evaluate the SRM directly, we use the final SURE-trained Qwen3-4B checkpoint as a fixed executor on all $600$ test problems, comprising $30$ AIME24, $30$ AIME25, $40$ AMC23, and $500$ MATH500 problems.
For each problem, we sample and score $20$ format-valid strategies and generate one binary-labeled answer per strategy.
The resulting $12{,}000$ strategies are partitioned into five within-problem score quantiles of $2{,}400$ strategies each.
The utility-ranking appendix provides the complete protocol.

\begin{table}[t]
\centering
\small
\caption{Downstream answer accuracy by within-problem SRM score quantile with final SURE-trained Qwen3-4B as the fixed executor.}
\label{tab:utility-ranking}
\setlength{\tabcolsep}{5pt}
\begin{tabular}{lrr}
\toprule
SRM quantile & Strategies & Accuracy (\%) \\
\midrule
80--100\% & 2,400 & 80.88 \\
60--80\%  & 2,400 & 78.76 \\
40--60\%  & 2,400 & 78.01 \\
20--40\%  & 2,400 & 77.63 \\
0--20\%   & 2,400 & 77.14 \\
\bottomrule
\end{tabular}

\end{table}

Accuracy increases strictly with the SRM quantile.
The top 20\% exceeds the bottom 20\% by $3.74\%$ and exceeds the second-highest quantile by $2.12\%$.
Because each strategy is executed once, this experiment evaluates relative group-level utility rather than calibration of individual SRM scores.
The monotonic trend shows that the SRM score contains meaningful information about the relative utility of strategies generated for the same problem (RQ1).
The ranking trend persists across SURE checkpoints.
On a shared Stage-1 strategy pool, frozen Qwen3-4B, Qwen3-8B, and Llama-3.2-3B Strategy-SFT executors yield pairwise ranking accuracies of $68.1\%$, $64.8\%$, and $64.6\%$, with Top-over-Bottom pass@1 gains of $3.5\%$, $3.0\%$, and $3.1\%$, respectively; full analyses appear in the appendix (Table~\ref{tab:cross-executor-utility}).

\subsection{SRM Data and Reward Design Ablations}
\label{sec:ablations}

\begin{figure}[t]
\centering
\includegraphics[width=\linewidth]{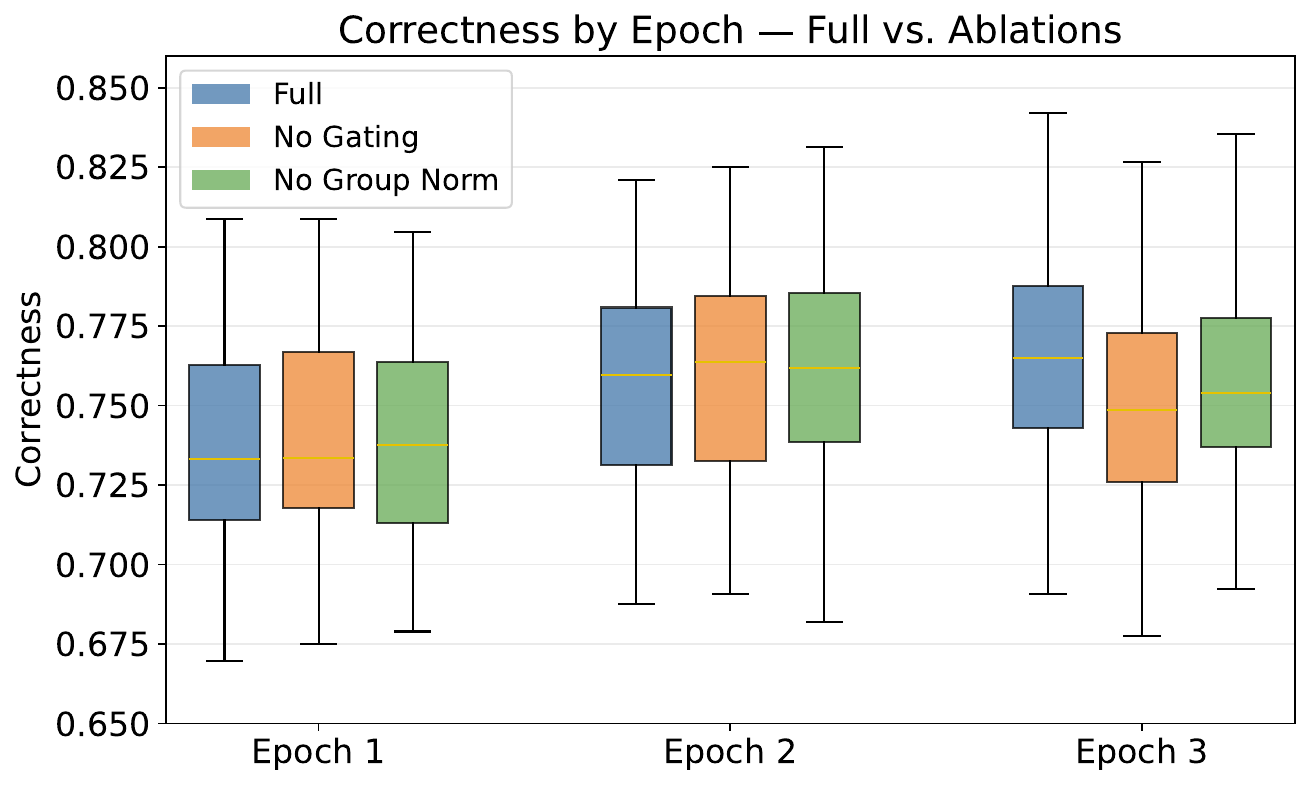}
\caption{Correctness by epoch on Qwen3-4B. Removing either stabilizer causes a plateau or regression in epoch 3.}
\label{fig:ablation-stability}

\end{figure}

We evaluate the two SRM preference sources and the two reward stabilizers by removing each component individually on Qwen3-4B.
Table~\ref{tab:ablation} reports final pass@1 after the full RL run.
Figure~\ref{fig:ablation-stability} tracks full SURE and the two reward-design ablations across the three RL epochs.

\begin{table}[t]
\centering
\small
\caption{SRM data and reward-design ablations on Qwen3-4B.}
\label{tab:ablation}
\setlength{\tabcolsep}{3pt}
\resizebox{\columnwidth}{!}{%
\begin{tabular}{lcccc|c}
\toprule
Variant & AIME24 & AIME25 & AMC23 & MATH500 & Avg \\
\midrule
\textbf{SURE} (full) & \textbf{42.67} & \textbf{31.00} & 84.25 & \textbf{83.02} & \textbf{60.24} \\
$-\mathcal{D}_{\text{rollout}}$ & 42.00 & 30.33 & 83.75 & 82.68 & 59.69 \\
$-\mathcal{D}_{\text{teacher}}$ & 42.33 & 30.33 & 84.00 & 82.94 & 59.90 \\
\midrule
$-$ correctness gating   & 40.67 & 30.33 & 83.75 & 82.24 & 59.25 \\
$-$ group normalization  & 39.33 & 30.67 & \textbf{84.50} & 82.54 & 59.26 \\
\bottomrule
\end{tabular}}

\end{table}

\paragraph{Both SRM data sources provide useful supervision.}
Removing rollout-induced and teacher-constructed pairs lowers Avg from $60.24\%$ to $59.69\%$ and $59.90\%$, respectively; both variants underperform full SURE on all four benchmarks.
The larger rollout-data drop indicates that execution-grounded preferences contribute more, while teacher pairs provide an additional gain.

\paragraph{Correctness gating contributes most on the hardest subset.}
This is consistent with reward-hacking analyses~\citep{gao2024designing}. Without the gate, the policy can drift toward strategies that the SRM scores favorably but that do not produce correct answers, an effect most visible on the hardest problems.
Figure~\ref{fig:ablation-stability} shows the temporal signature of this drift.
The No-Gating variant reaches its highest accuracy at epoch 2 but \emph{regresses} by epoch 3, while the gated configuration avoids this late decline.
This temporal pattern is consistent with reward hacking and indicates that gating suppresses utility for incorrect responses while preserving SRM ranking among correct ones, a favorable tradeoff.

\paragraph{Group-wise normalization protects against late-training drift.}
Removing normalization decreases Avg by $0.98\%$, comparable to the gating ablation.
Figure~\ref{fig:ablation-stability} reveals a more specific pattern.
The No-Group-Norm variant tracks the full configuration through epoch 2 and only diverges in epoch 3, with the median falling below the full curve.
Score-scale drift across rollouts within a group accumulates as RL proceeds, eventually pushing the policy toward strategies whose SRM scores are inflated relative to peers.
Group normalization is a cheap protection against this late-training failure mode.

\subsection{GRPO-Stage Cost Given a Pre-Trained SRM (RQ3)}
\label{sec:compute-matched}

PTA-style methods estimate plan quality online during RL by repeatedly resampling executions from the current policy~\citep{dou2025plan}.
SURE moves this estimation to a pretrained reward model that is frozen before RL begins.
We measure GRPO-stage GPU cost for each method.
For SURE, this includes GRPO training and frozen SRM inference during rollout scoring.
For PTA-GRPO, it includes GRPO training and per-step online resampling.
For GRPO+PRM, it includes GRPO training and PRM inference.
Constructing the SRM preference data took 47 wall-clock hours on a single node with eight NVIDIA RTX 4090 GPUs.
This one-time offline preparation cost is separate from the GRPO-stage comparison, and we will release the complete preference data for reuse in future and related work.

\begin{table}[t]
\centering
\small
\caption{GRPO-stage cost on Qwen3-4B. SURE adds about 6 GPU-h over outcome-only GRPO and uses $45\%$ less compute than GRPO+PRM.}
\label{tab:compute-matched}
\setlength{\tabcolsep}{3pt}
\resizebox{\columnwidth}{!}{%
\begin{tabular}{lcccc}
\toprule
Method & GRPO-stage GPU-h & AMC23 & MATH500 \\
\midrule
GRPO (outcome+format)              & 168.06 & 80.50 & 82.32 \\
PTA-GRPO                           & 235.29 & 82.25 & 81.54 \\
GRPO + PRM                         & 318.30 & 82.00 & 82.34 \\
\textbf{SURE}                      & 174.05 & 84.25 & 83.02 \\
\bottomrule
\end{tabular}}

\end{table}

SURE outperforms PTA-GRPO by $+2.00$ on AMC23 and $+1.48$ on MATH500 while running $61$ GPU-hours faster at the GRPO stage.
Compared with GRPO+PRM, SURE achieves comparable accuracy ($+2.25$ AMC23, $+0.68$ MATH500) with $45\%$ less GRPO-stage compute (174.05 vs.\ 318.30 GPU-h).

\section{Conclusion}
\label{sec:conclusion}
We presented SURE, which trains a Strategy Reward Model on confidence-aware pairwise preferences and uses it during policy optimization.
With correctness gating and group-wise normalization, SURE improves over outcome-only GRPO across three backbones, integrates with step-level PRMs, and achieves competitive accuracy with substantially lower GRPO-stage compute than online plan-reward baselines.
The SRM also transfers as a standalone strategy selector, indicating that it captures reusable strategy quality.
These findings show that explicitly evaluating high-level decisions can improve reasoning policies beyond terminal correctness alone.
By separating offline strategy evaluation from policy optimization, SURE provides a practical way to reuse strategic supervision across training runs and model backbones.
\FloatBarrier
\newpage
\begingroup
\small
\setlength{\bibsep}{2pt}
\bibliographystyle{abbrvnat}
\bibliography{references}
\endgroup

\clearpage
\appendix
\label{sec:appendix}
\section{Comparison with Online Plan-Reward}
\label{sec:appendix-comparison}

SURE differs structurally from methods that estimate plan quality online during RL~\citep{dou2025plan}.
The comparison has three aspects.

\textbf{Offline vs.\ online estimation.}
These methods estimate plan quality from rollouts of the \emph{current} policy throughout training, requiring repeated sampling and quality re-estimation as the policy evolves.
The SRM is trained once from rollouts of a fixed SFT policy and reused without modification throughout GRPO.
This moves strategy utility estimation from a recurring online computation to a one-time offline stage.
The two-level rollout procedure is performed once rather than repeated on each RL update.

\textbf{Unit of credit.}
Such methods typically evaluate plan and action trajectories holistically, conflating plan choice with execution.
The SRM reads the strategy text alone and never observes execution tokens or the final answer.
Its score is then combined with correctness and format rewards at the sequence level.

\textbf{GRPO-stage cost.}
We measure the cost of the GRPO stage itself, assuming a strategy evaluator is already available.
For SURE, the only overhead over outcome-only GRPO is a frozen SRM inference call per rollout ($\sim$6 GPU-h).
For PTA-GRPO, the overhead is per-step online resampling that recurs throughout training.
This framing conditions the comparison on a pretrained SRM and reports only the GRPO-stage cost.
SRM training is a one-time offline investment that is reused across runs, backbones, and tasks.

The SRM occupies a specific design point that is offline, strategy-focused, and grounded in execution evidence.

\section{SFT Strategy Generation Prompt}
\label{sec:appendix-sft-prompt}

For Stage 1 SFT, we prompt Qwen3.5-397B-A17B with each problem and its reference chain-of-thought solution to distill a high-level strategy.
The teacher is instructed to extract the overall approach, key theorems, and major branching decisions while excluding numerical computations, low-level derivations, and the final answer.
The resulting (question, strategy, execution) triples are then manually verified before being added to the SFT corpus.
The prompt template is shown in Figure~\ref{fig:sft-prompt}.

\section{SRM Preference Data Construction Details}
\label{sec:appendix-srm-data}

\subsection{Data Pipeline and Statistics}

We begin with $32{,}000$ problems sampled from DeepMath and OpenThoughts.
For each problem, we sample $K\!=\!8$ candidate strategies from $\pi_\text{SFT}$ and generate $M\!=\!24$ independent execution rollouts per strategy, yielding $192$ rollouts per problem.
After computing empirical success rates and applying the Beta-Binomial confidence filter ($\mathbb{P}[q(s^+)>q(s^-)]\geq 0.9$), $53{,}464$ preference pairs survive.
These are then passed through the LLM judge consistency check (described below), and $48{,}413$ pairs are retained after discarding judge--rollout disagreements.
The teacher-constructed contrastive set contributes an additional $17{,}197$ pairs.
After deduplication and format filtering, the merged training set $\mathcal{D}_\text{SRM}$ contains $\mathbf{65{,}610}$ preference pairs (Table~\ref{tab:srm-data-stats}).

\begin{table}[h]
\centering
\small
\caption{SRM preference data statistics. Rollout-induced pairs are constructed from 32,000 problems (DeepMath + OpenThoughts) with $K\!=\!8$ strategies and $M\!=\!24$ rollouts each.}
\label{tab:srm-data-stats}
\resizebox{\columnwidth}{!}{
\begin{tabular}{lrr}
\toprule
Source & Raw pairs & Retained \\
\midrule
Rollout-induced (post confidence filter)  & 53,464 & --- \\
\quad + LLM judge consistency filter      & ---    & 48,413 \\
Teacher-constructed                       & 17,197 & 17,197 \\
\midrule
\textbf{Total} $\mathcal{D}_\text{SRM}$  & ---    & \textbf{65,610} \\
\bottomrule
\end{tabular}}
\end{table}

Algorithm~\ref{alg:srm-data} formalizes the construction pipeline.

\begin{algorithm}[h]
\small
\caption{SRM Preference Data Construction}
\label{alg:srm-data}
\begin{algorithmic}[1]
\Require SFT policy $\pi_{\text{SFT}}$, problems $\mathcal{D}_{\text{prob}}$, judge $\mathcal{J}$, teacher $\mathcal{T}$, group sizes $K, M$, confidence threshold $\tau$
\State $\mathcal{D}_{\text{rollout}} \leftarrow \emptyset$,\ \ $\mathcal{D}_{\text{teacher}} \leftarrow \emptyset$
\For{each $x \in \mathcal{D}_{\text{prob}}$} \Comment{rollout-induced source}
  \State Sample $\{s^{(k)}\}_{k=1}^{K} \sim \pi_{\text{SFT}}(\cdot \mid x)$
  \For{$k = 1, \ldots, K$}
    \State Sample $M$ executions conditioned on $s^{(k)}$; $c^{(k)} \leftarrow$ \#correct
    \State $\hat q^{(k)} \leftarrow c^{(k)}/M$
  \EndFor
  \For{each $(i, j)$ with $\hat q^{(i)} > \hat q^{(j)}$}
    \State $p_{ij} \leftarrow \mathbb{P}\bigl[q^{(i)} > q^{(j)} \mid c^{(i)}, c^{(j)}, M\bigr]$ \Comment{Beta-Binomial, Eq.~\ref{eq:confidence}}
    \If{$p_{ij} \geq \tau$ \textbf{and} $\mathcal{J}(x, s^{(i)}, s^{(j)})\!=\!\textsc{Reasonable}$}
      \State $\mathcal{D}_{\text{rollout}} \leftarrow \mathcal{D}_{\text{rollout}} \cup \{(x, s^{(i)}, s^{(j)})\}$
    \EndIf
  \EndFor
\EndFor
\For{each $x \in \mathcal{D}_{\text{prob}}$} \Comment{teacher-constructed source}
  \State $(s^+, s^-) \leftarrow \mathcal{T}(x)$
  \If{$s^+, s^-$ both pass deterministic format check}
    \State $\mathcal{D}_{\text{teacher}} \leftarrow \mathcal{D}_{\text{teacher}} \cup \{(x, s^+, s^-)\}$
  \EndIf
\EndFor
\State $\mathcal{D}_{\text{SRM}} \leftarrow \textsc{Dedup}(\mathcal{D}_{\text{rollout}} \cup \mathcal{D}_{\text{teacher}})$
\State \Return $\mathcal{D}_{\text{SRM}}$
\end{algorithmic}
\end{algorithm}

\subsection{Strategy Extraction and Deterministic Tag Checking}
\label{sec:appendix-tag-parser}

We use a deterministic parser to extract the \texttt{<strategy>} and \texttt{<think>} segments from each model output.
The parser is responsible only for structural validation and block extraction.
It checks whether the output contains exactly one \texttt{<strategy>}\dots\texttt{</strategy>} block, followed by exactly one \texttt{<think>}\dots\texttt{</think>} block, followed by a \texttt{\textbackslash boxed\{\}} expression containing the final answer.
Format-invalid outputs are filtered out before SRM training and reward computation.
The LLM judge described below is \emph{not} used for tag checking or answer extraction. All tag validation is performed by deterministic rules.

\subsection{LLM-Assisted Preference Reasonableness Verification}
\label{sec:appendix-judge-prompt}

To reduce noisy or spurious preference labels, we use an LLM judge (Qwen3.5-397B-A17B) to verify whether a constructed strategy preference is reasonable, following the LLM-as-a-Judge paradigm~\citep{zheng2023llmjudge}.
Unlike a pairwise reranker that independently picks between two unlabeled strategies, our judge is given the proposed preference label directly.
For each preference pair $(x, s^{+}, s^{-})$, the judge receives the problem $x$, the preferred strategy $s^{+}$, and the rejected strategy $s^{-}$, and decides whether $s^{+} \succ s^{-}$ is justified.
The judge does not observe execution rollouts, empirical success rates, or final answers. It evaluates only whether the preferred strategy is mathematically more sound and more likely to lead to a correct solution.

For rollout-induced pairs, $s^{+}$ is the strategy with higher estimated execution success after confidence-aware filtering.
A pair is retained only if the judge outputs \texttt{REASONABLE}. Pairs labeled \texttt{UNREASONABLE}, with invalid labels, or with ambiguous responses are discarded.
The prompt template is shown in Figure~\ref{fig:judge-prompt}.

\subsection{Teacher-Constructed Contrastive Prompt}
\label{sec:appendix-teacher-prompt}

Because rollout-induced pairs are limited to strategies the SFT policy can generate, we also use Qwen3.5-397B-A17B to construct positive and negative strategy pairs for each problem.
The prompt template is shown in Figure~\ref{fig:teacher-prompt}.

\noindent Generated pairs are subject to the same format-validity check applied to rollout-induced pairs.
Teacher-constructed negatives provide contrastive signal for recognizable structural errors (wrong theorem, infeasible subgoal order) that may be underrepresented in rollout-induced pairs, where both strategies must be producible by the SFT policy.

\section{Implementation Details}
\label{sec:appendix-impl}

All experiments are run on 4$\times$A100 80GB GPUs in bfloat16 precision.
Table~\ref{tab:training_hyperparams} lists the full hyperparameter configuration for the three stages.
Below we summarize the role of each stage.

\paragraph{Stage 1 Strategy-Guided SFT.}
The base policy (Qwen3-4B) is fully fine-tuned with FSDP2 to learn the structured \texttt{<strategy>}/\texttt{<think>} output format using the standard next-token prediction loss.
This stage establishes the structured output format rather than improving raw reasoning ability.

\paragraph{Stage 2 SRM Training.}
The SRM is initialized from Qwen2.5-7B-Instruct, adapted with LoRA, and trained with the margin-augmented Bradley--Terry loss with scale anchoring (Eq.~\ref{eq:srm-loss}).
Training takes approximately $4.38$ hours on the 4$\times$A100 80GB setup.
The preference data are constructed offline in 47 wall-clock hours on a single node with eight NVIDIA RTX 4090 GPUs.
The trained SRM is frozen before Stage 3 and is shared across all GRPO experiments. Its training cost is not attributed to any individual run in the compute comparison.

\paragraph{Stage 3 SURE Policy Optimization.}
GRPO is initialized from the Stage 1 SFT checkpoint and trained with full parameters under DeepSpeed ZeRO-2.
Each rollout receives a composite reward combining correctness, format, and the frozen SRM score, with group-wise normalization and correctness gating. KL regularization follows~\citet{shao2024deepseekmath}.

\paragraph{Inference and evaluation.}
All evaluation rollouts are sampled with temperature $0.6$, top-$p$ $0.95$, and maximum new tokens $32{,}768$ (matching the maximum sequence length, $32{,}768$).
We use a maximum batch size of $32$ for inference.
For each problem we draw 10 independent samples and report the mean pass@1.
Final answers are extracted by a deterministic parser from the \texttt{\textbackslash boxed\{\}} expression that follows the \texttt{</think>} block, and compared against the gold answer using a math-aware verifier.

\begin{table*}[t]
\centering
\small
\caption{Training hyperparameters for the three-stage SURE pipeline on the Qwen3-4B backbone. The same configuration is applied to the Qwen3-8B and Llama-3.2-3B-Instruct backbones, except for the policy model and the matching Stage 1 SFT checkpoint used to initialize Stage 3.}
\label{tab:training_hyperparams}
\begin{tabular}{lccc}
\hline
\textbf{Hyperparameter} & \textbf{Stage 1 SFT} & \textbf{Stage 2 SRM} & \textbf{Stage 3 SURE} \\
\hline
\multicolumn{4}{l}{\textit{Model}} \\
Base / policy model & Qwen3-4B & Qwen2.5-7B-Instruct & Stage-1 checkpoint \\
Reward model & -- & -- & Stage-2 SRM (frozen) \\
Training type & Full fine-tuning & LoRA & Full fine-tuning \\
LoRA rank / $\alpha$ / target & -- & 64 / 128 / all-linear & -- \\
Reward / value head & -- & Scalar value head & -- \\
\hline
\multicolumn{4}{l}{\textit{Data}} \\
Training data size & 17{,}467 triples & 65{,}610 pairs & 7{,}200 problems \\
\hline
\multicolumn{4}{l}{\textit{Optimization}} \\
Number of epochs & 3 & 3 & 3 \\
Learning rate & $4\times10^{-5}$ & $1\times10^{-4}$ & $1\times10^{-6}$ \\
Warmup ratio & 0.05 & 0.05 & -- \\
Per-device batch size & 1 & 1 & 2 \\
Gradient accumulation & 8 & 4 & 16 \\
Effective global batch & 32 & 16 & 128 \\
Number of GPUs & 4$\times$A100 80GB & 4$\times$A100 80GB & 4$\times$A100 80GB \\
Precision & bfloat16 & bfloat16 & bfloat16 \\
Distributed strategy & FSDP2 & DeepSpeed ZeRO-2 & DeepSpeed ZeRO-2 \\
Gradient checkpointing & Enabled & -- & -- \\
\hline
\multicolumn{4}{l}{\textit{Sequence length}} \\
Maximum input length & 8{,}192 & 2{,}048 & 4{,}095 \\
Maximum completion length & -- & -- & 4{,}096 \\
\hline
\multicolumn{4}{l}{\textit{Loss / Reward}} \\
Loss function & Next-token prediction & Margin BT + scale anchor & GRPO clipped surrogate \\
Margin $m$ & -- & 0.5 & -- \\
Scale regularization $\lambda$ & -- & 0.02 & -- \\
Reward components & -- & -- & Correctness, format, SRM \\
Reward weights ($\alpha / \beta / \gamma$) & -- & -- & 0.7 / 0.1 / 0.2 \\
\hline
\multicolumn{4}{l}{\textit{GRPO-specific}} \\
GRPO group size $G$ & -- & -- & 8 \\
Sampling temperature & -- & -- & 0.9 \\
KL coefficient & -- & -- & 0.02 \\
vLLM mode / GPU mem util & -- & -- & Colocate / 0.35 \\
\hline
\end{tabular}
\end{table*}

\section{SFT Teacher Strength Ablation Details}
\label{sec:appendix-teacher-ablation}

The teacher-strength ablation changes only the model used to construct the Stage 1 SFT strategy targets on Qwen3-4B.
The strong setting uses Qwen3.5-397B-A17B, while the ablation setting uses Qwen3-4B to generate the same strategy-guided format.
The SFT data size, policy optimization schedule, SRM preference data, SRM parameters, and LLM judge are unchanged.
Consequently, the comparison isolates the effect of SFT strategy teacher strength rather than changing the learned utility signal.

\begin{table}[h]
\centering
\small
\caption{SFT teacher strength ablation details on Qwen3-4B. Only the teacher used for Stage 1 strategy construction is changed.}
\label{tab:appendix-teacher-ablation}
\setlength{\tabcolsep}{3pt}
\resizebox{\columnwidth}{!}{%
\begin{tabular}{lcccc|c}
\toprule
SFT strategy teacher & AIME24 & AIME25 & AMC23 & MATH500 & Avg \\
\midrule
Qwen3.5-397B-A17B & 42.67 & 31.00 & \textbf{84.25} & \textbf{83.02} & \textbf{60.24} \\
Qwen3-4B & 42.67 & 31.00 & 82.25 & 82.70 & 59.66 \\
Difference & 0.00 & 0.00 & $-2.00$ & $-0.32$ & $-0.58$ \\
\bottomrule
\end{tabular}}
\end{table}

The weaker SFT teacher lowers the average by only $0.58\%$.
The result supports the interpretation that SURE's final gains are not attributable solely to distillation from the largest teacher during SFT.

\section{SURE Policy Optimization Loop}
\label{sec:appendix-algorithm}

Algorithm~\ref{alg:sure-rl} summarizes the SURE policy optimization update.
For each problem, $G$ rollouts are sampled from the current policy.
A deterministic parser extracts the strategy and execution blocks, the verifier checks answer correctness, and the frozen SRM scores the strategy text.
SRM scores are normalized within the group, gated by correctness, and combined with format and correctness rewards into a composite scalar that drives the standard GRPO update.

\begin{algorithm}[h]
\small
\caption{SURE Policy Optimization}
\label{alg:sure-rl}
\begin{algorithmic}[1]
\Require SFT policy $\pi_{\text{SFT}}$, frozen SRM $r_\phi$, problems $\mathcal{D}_{\text{RL}}$, group size $G$, coefficients $\alpha, \beta, \gamma$
\State $\pi_\theta \leftarrow \pi_{\text{SFT}}$
\For{each training step}
  \State Sample $x \sim \mathcal{D}_{\text{RL}}$
  \State Sample group $\{o_g\}_{g=1}^{G} \sim \pi_\theta(\cdot \mid x)$
  \For{$g = 1, \ldots, G$}
    \State $(s_g, y_g, \text{ok}_g) \leftarrow \text{ParseTags}(o_g)$
    \State $R^{(g)}_{\text{format}} \leftarrow \mathbf{1}[\text{ok}_g]$
    \State $R^{(g)}_{\text{correct}} \leftarrow \mathbf{1}[\text{verify}(y_g)]$
    \State $R^{(g)}_{\text{SRM}} \leftarrow r_\phi(x, s_g)$ \Comment{strategy only}
  \EndFor
  \State $\tilde R^{(g)}_{\text{SRM}} \leftarrow \text{GroupNorm}\bigl(\{R^{(g)}_{\text{SRM}}\}_{g=1}^G\bigr)$
  \For{$g = 1, \ldots, G$}
    \State $\hat R^{(g)}_{\text{SRM}} \leftarrow R^{(g)}_{\text{correct}} \cdot \tilde R^{(g)}_{\text{SRM}}$ \Comment{gating}
    \State $R^{(g)} \leftarrow \alpha R^{(g)}_{\text{correct}} + \beta R^{(g)}_{\text{format}} + \gamma \hat R^{(g)}_{\text{SRM}}$
  \EndFor
  \State Compute group-relative advantages $\hat A^{(g)}$ from $\{R^{(g)}\}$
  \State Update $\pi_\theta$ via clipped surrogate~\citep{schulman2017ppo} with KL penalty toward $\pi_{\text{SFT}}$
\EndFor
\State \Return $\pi_\theta$
\end{algorithmic}
\end{algorithm}

\section{SRM Utility Ranking Details}
\label{sec:appendix-utility-ranking}

We evaluate SRM ranking on $600$ held-out problems.
The fixed executor is the final SURE-trained Qwen3-4B policy checkpoint.
The evaluation pool contains $30$ AIME24, $30$ AIME25, $40$ AMC23, and $500$ MATH500 problems, matching the four test sets used elsewhere in the paper.
For each problem, we sample $20$ candidate strategies and generate one answer for each strategy with the same fixed executor.
This produces $12{,}000$ strategy and answer pairs.
All strategies contain exactly one non-empty \texttt{<strategy>} block, and format validation removes none of them.
Within each problem, we sort strategies by SRM score and divide them into five quantiles with four strategies per quantile.
The aggregated quantiles each contain $2{,}400$ strategies.
Because each strategy receives only one binary answer label, the analysis evaluates relative group-level utility and does not claim calibration of individual SRM scores.

\newpage
\section{SRM Ranking Across SURE Checkpoints}
\label{sec:appendix-checkpoint-ranking}

We repeat the SRM ranking study at the three SURE training checkpoints.
At each epoch, candidate strategies are regenerated by the corresponding Qwen3-4B checkpoint, so the analysis captures the combined effect of executor evolution and the changing policy-induced strategy distribution.
The evaluation uses the same $600$ test problems, with $20$ candidate strategies and one downstream answer per strategy.

\begin{table}[t]
\centering
\small
\setlength{\tabcolsep}{5pt}
\caption{Downstream answer accuracy by SRM score quantile across SURE checkpoints. Candidate strategies are regenerated by the corresponding checkpoint at each epoch, and each strategy receives one downstream execution.}
\label{tab:checkpoint-ranking}
\begin{tabular}{lccc}
\toprule
SRM quantile & Epoch 1 & Epoch 2 & Epoch 3 \\
\midrule
80--100\% & 80.20 & 80.78 & 80.88 \\
60--80\%  & 78.70 & 78.80 & 78.76 \\
40--60\%  & 77.55 & 77.85 & 78.01 \\
20--40\%  & 77.33 & 77.60 & 77.63 \\
0--20\%   & 76.96 & 77.21 & 77.14 \\
\bottomrule
\end{tabular}
\end{table}

The quantile ordering is strictly monotonic at every checkpoint.
The top-to-bottom accuracy gap is $3.24\%$, $3.57\%$, and $3.74\%$ at epochs 1, 2, and 3, respectively.
Thus, the ranking signal does not show degradation under the evolving policy-induced strategy distribution, although this analysis does not isolate executor drift from strategy-distribution drift.


\section{Cross-Executor Strategy Utility Validation}
\label{sec:appendix-cross-executor}

We evaluate cross-executor transfer on $300$ evaluation problems.
For each problem, the Qwen3-4B Stage-1 Strategy-SFT policy samples $K=8$ candidate strategies, producing an independent pool of $2{,}400$ strategies that has not been influenced by SRM-guided policy optimization.
The same candidate pool is used for all three frozen Strategy-SFT executors: Qwen3-4B, Qwen3-8B, and Llama-3.2-3B-Instruct.
For each strategy, each executor generates $M=8$ independent execution rollouts.
This yields $300 \times 8 \times 8 = 19{,}200$ executions per executor.
Neither the SRM nor any executor is updated during evaluation.

For executor $e$, we estimate the utility of strategy $s^{(k)}$ by its empirical success rate over the eight conditioned executions.
For within-problem strategy pairs with different empirical success rates, Pairwise Acc. measures whether the ordering induced by the SRM scores agrees with the executor-specific empirical utility ordering.
Top and Bottom denote the strategies with the highest and lowest SRM scores for each problem, respectively.
Their reported accuracies are averaged over the eight conditioned executions and all evaluation problems.

\begin{table}[t]
\centering
\small
\caption{Cross-executor strategy utility validation with a shared strategy pool generated by the Qwen3-4B Stage-1 Strategy-SFT policy. Pairwise Acc. measures agreement with executor-specific empirical utility rankings within each problem. All values are percentages.}
\label{tab:cross-executor-utility}
\setlength{\tabcolsep}{3pt}
\resizebox{\columnwidth}{!}{%
\begin{tabular}{lrrrr}
\toprule
Executor & Pairwise Acc. & Bottom & Top & Top$-$Bottom \\
\midrule
Qwen3-4B Strategy-SFT & 68.1 & 55.5 & 59.0 & +3.5 \\
Qwen3-8B Strategy-SFT & 64.8 & 56.4 & 59.4 & +3.0 \\
Llama-3.2-3B Strategy-SFT & 64.6 & 17.7 & 20.8 & +3.1 \\
\bottomrule
\end{tabular}}
\end{table}

Pairwise ranking accuracy is above the $50\%$ chance level for every executor.
Selecting the highest-scoring strategy instead of the lowest-scoring strategy increases pass@1 from $55.5\%$ to $59.0\%$ for Qwen3-4B, from $56.4\%$ to $59.4\%$ for Qwen3-8B, and from $17.7\%$ to $20.8\%$ for Llama-3.2-3B.
The results show that the SRM ranking signal transfers to a larger Qwen executor and to the Llama model family.
The highest ranking agreement occurs for Qwen3-4B, the executor used to construct the rollout-induced preferences.
This pattern is consistent with strategy utility being executor-conditioned rather than an intrinsic property of the strategy text, while demonstrating that the learned signal remains partially transferable.

\section{Qualitative Example}
\label{sec:appendix-example}

\begin{figure*}[h]
\centering
\includegraphics[width=\textwidth,height=0.82\textheight,keepaspectratio]{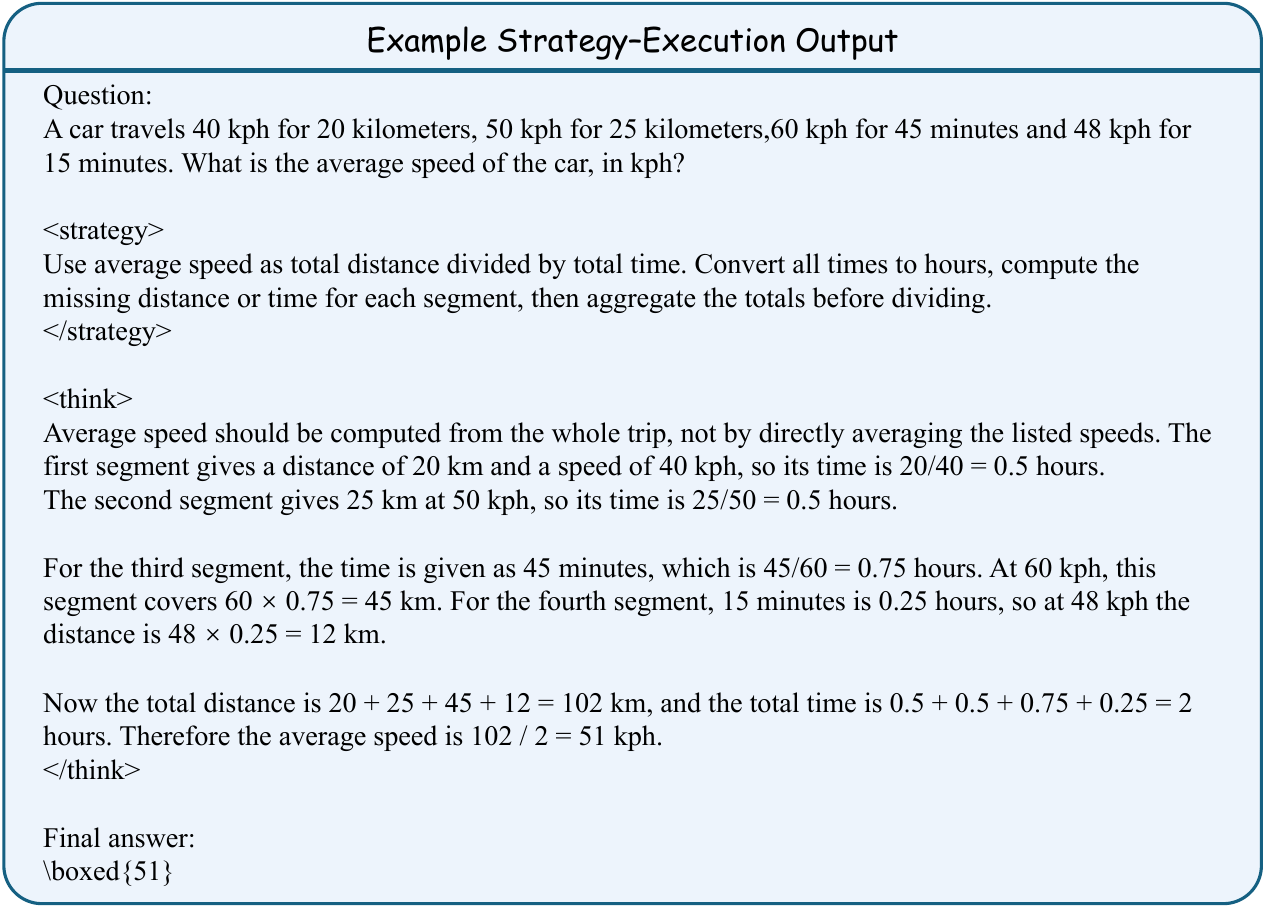}
\caption{A complete (question, strategy, execution) example. The \texttt{<strategy>} block specifies the high-level route, while the \texttt{<think>} block performs the detailed derivation that produces the final answer.}
\label{fig:example-output}
\end{figure*}

To illustrate what the strategy/execution decomposition looks like in practice, Figure~\ref{fig:example-output} shows a complete (question, strategy, execution) instance from our SFT corpus.
The strategy block captures the high-level approach, including theorem choice, key transformations, and subgoal ordering, without numerical computation.
The execution block then carries out the concrete derivation under the strategy prefix.

\section{Dataset Licenses and Ethical Considerations}
\label{sec:appendix-ethics}

\paragraph{Datasets.}
We use three publicly released mathematical reasoning datasets within their intended research use.
MATH~\citep{hendrycks2021math} is released under the MIT License;
OpenThoughts~\citep{guha2025openthoughts} is released under the Apache 2.0 License;
DeepMath-103K~\citep{he2025deepmath} is released under the MIT License.
For evaluation, AIME 2024, AIME 2025, and AMC 2023 problems are publicly released competition problems. MATH500 is the standard test split of MATH used by~\citet{lightman2023let}.
SFT uses MATH-train, OpenThoughts, and DeepMath, sampled after deduplication against AIME 2024, AIME 2025, AMC 2023, and MATH500. The subset drawn from MATH uses only the training split, and MATH500 is never used in SFT, SRM training, or RL.

\paragraph{Models.}
The Qwen3-4B/8B, Qwen2.5-7B-Instruct, and Qwen2.5-Math-PRM-7B models~\citep{yang2025qwen3,yang2024qwen25,yang2024qwen25math} are released under the Apache 2.0 License.
Llama-3.2-3B-Instruct is released under the Meta Llama 3 Community License.
The Qwen3.5-397B-A17B teacher model is accessed via its publicly available API for strategy generation and judge filtering, and we follow the provider's terms of service.
All released model checkpoints in this work will be distributed under the Apache 2.0 License, consistent with the policies of the upstream models.

\begin{figure*}[h]
\centering
\includegraphics[width=\textwidth,height=0.82\textheight,keepaspectratio]{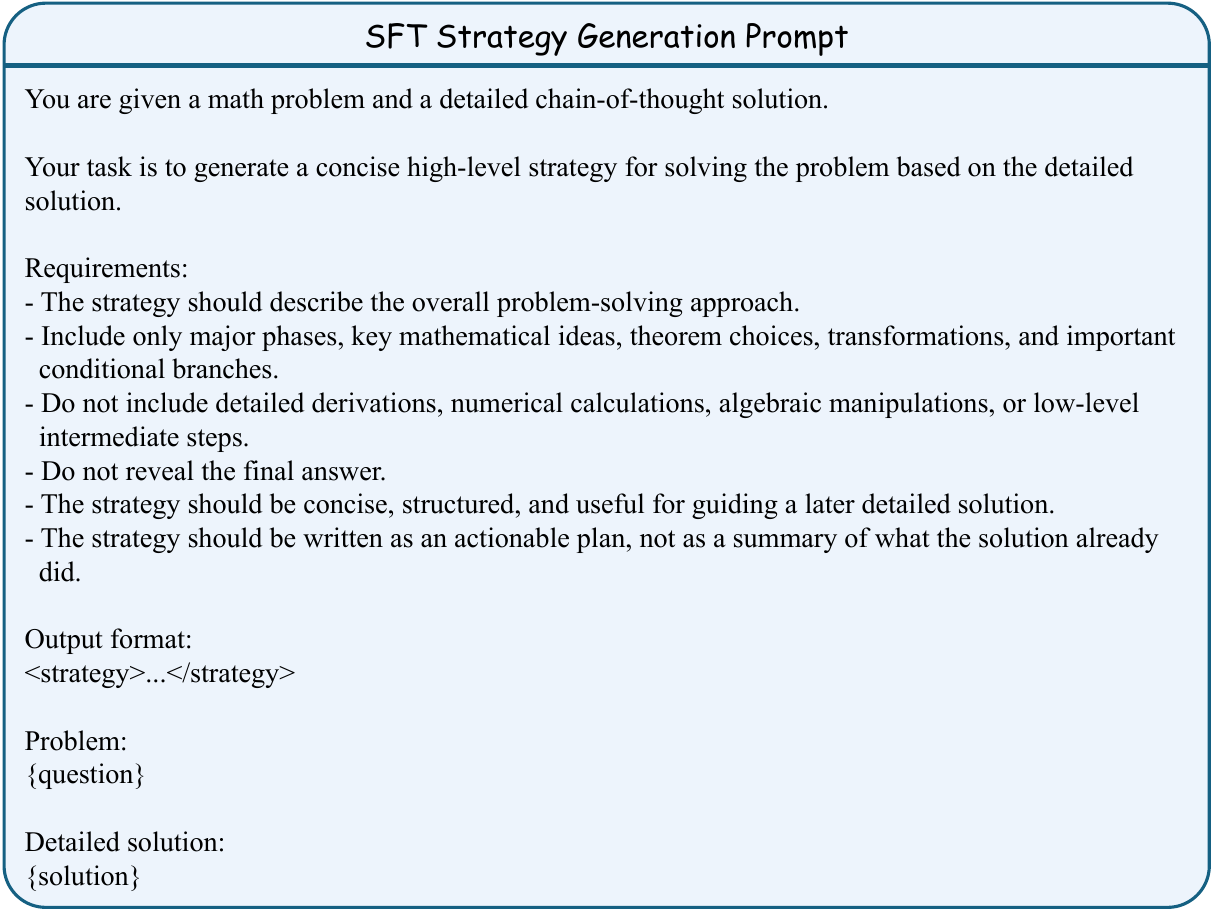}
\caption{SFT strategy generation prompt used to distill high-level strategies from teacher chain-of-thought solutions.}
\label{fig:sft-prompt}
\end{figure*}

\begin{figure*}[h]
\centering
\includegraphics[width=\textwidth,height=0.82\textheight,keepaspectratio]{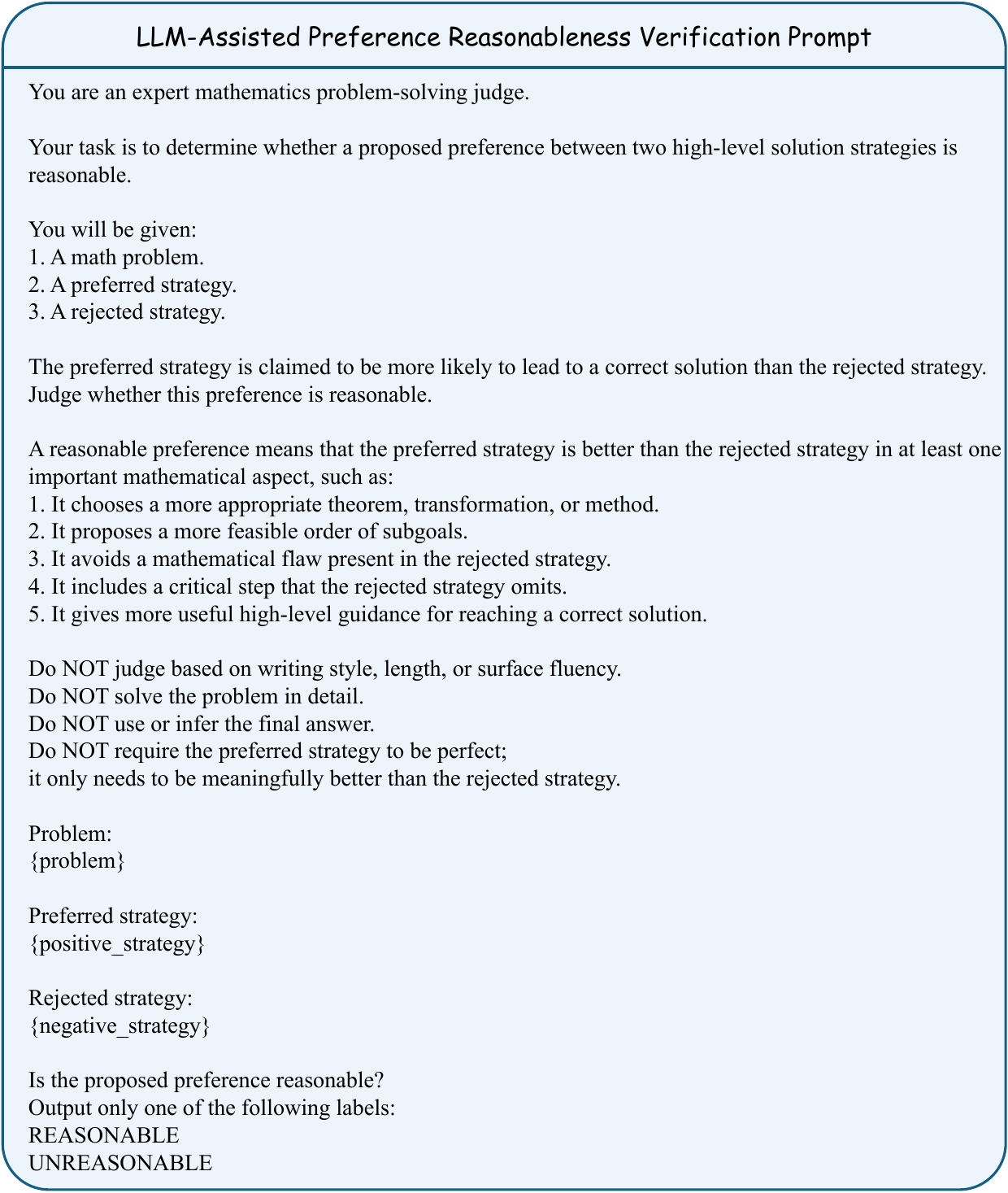}
\caption{LLM-assisted preference reasonableness verification prompt. Given a problem and a labeled preference $(s^+, s^-)$, the judge decides whether the proposed preference is justified.}
\label{fig:judge-prompt}
\end{figure*}

\begin{figure*}[h]
\centering
\includegraphics[width=\textwidth,height=0.82\textheight,keepaspectratio]{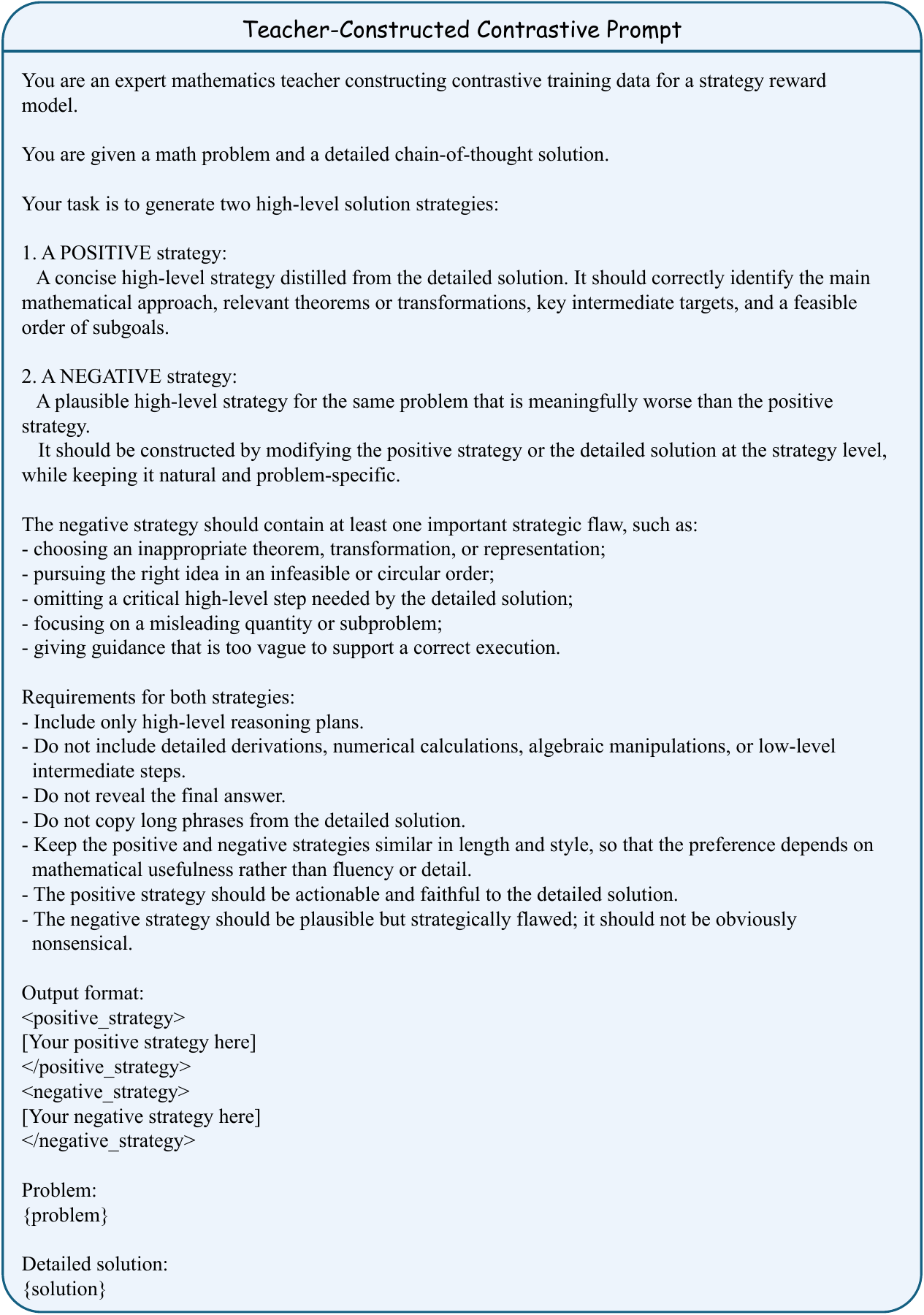}
\caption{Teacher-constructed contrastive prompt. The teacher LLM generates a paired positive and negative strategy for each problem, exposing recognizable structural errors that may be underrepresented in rollout-induced data.}
\label{fig:teacher-prompt}
\end{figure*}

The full prompt templates referenced in earlier sections are collected here.
Figure~\ref{fig:sft-prompt} shows the SFT strategy generation prompt; Figure~\ref{fig:judge-prompt} shows the LLM-assisted preference verification prompt; Figure~\ref{fig:teacher-prompt} shows the teacher-constructed contrastive prompt.

\end{document}